\DocumentMetadata{}
\RequirePackage{pdfmanagement-testphase}

\documentclass[sigconf]{acmart}

\usepackage{tabularx}
\usepackage{graphicx}
\usepackage{subcaption}
\usepackage{caption}
\usepackage{algorithm}
\usepackage{algorithmic}
\usepackage{url}
\usepackage{hyperref}
\usepackage{tikz}
\usepackage[most]{tcolorbox}
\usepackage{booktabs}

\definecolor{darkblue}{HTML}{00205b}
\definecolor{blue}{HTML}{a5b0cb}
\definecolor{teal}{HTML}{d4e6e8}
\definecolor{lightorange}{HTML}{eabcad}
\definecolor{orange}{HTML}{d58570}

\AtBeginDocument{%
  }

\setcopyright{none}
\renewcommand\footnotetextcopyrightpermission[1]{}
\begin{document}

\title{MedEqualizer: A Framework Investigating Bias in Synthetic Medical Data and Mitigation via Augmentation}






\author{Sama Salarian}
\authornote{Both authors contributed equally to this research.}
\email{salarian.1@osu.edu}

\author{Yue Zhang}
\authornotemark[1]
\email{zhang.8016@osu.edu}
\affiliation{%
  \institution{The Ohio State University}
  \city{Columbus}
  \country{USA}
}

\author{Swati Padhee}
\affiliation{%
  \institution{The Ohio State University}
  \city{Columbus}
  \country{USA}
}
\email{padhee.3@osu.edu}

\author{Srinivasan Parthasarathy}
\affiliation{%
  \institution{The Ohio State University}
  \city{Columbus}
  \country{USA}
}
\email{srini@cse.ohio-state.edu}


\begin{abstract}
Synthetic healthcare data generation presents a viable approach to enhance data accessibility and support research by overcoming limitations associated with real-world medical datasets. However, ensuring fairness across protected attributes in synthetic data is critical to avoid biased or misleading results in clinical research and decision-making. In this study, we assess the fairness of synthetic data generated by multiple generative adversarial network (GAN)-based models using the MIMIC-III dataset, with a focus on representativeness across protected demographic attributes. We measure subgroup representation using the logarithmic disparity metric and observe significant imbalances, with many subgroups either underrepresented or overrepresented in the synthetic data, compared to the real data. To mitigate these disparities, we introduce MedEqualizer, a model-agnostic augmentation framework that enriches the underrepresented subgroups prior to synthetic data generation.
Our results show that MedEqualizer significantly improves demographic balance in the resulting synthetic datasets, offering a viable path towards more equitable and representative healthcare data synthesis.
\end{abstract}




\maketitle

\section{Introduction}
Medical data records are generated by millions of individuals every day, spanning routine check-ups to complex surgical procedures. These records contain valuable information that can be harnessed to improve healthcare outcomes, advance medical research, and develop new treatments \citep{jensen2012mining}. For instance, electronic health records (EHRs) provide insights into disease progression, treatment efficacy, and patient outcomes \citep{gunasekar2016phenotyping}.

Despite their potential, access to medical data is often limited due to stringent data sharing regulations and institutional constraints. In the United States, regulations such as the Health Insurance Portability and Accountability Act (HIPAA) restrict the use and dissemination of health information, posing challenges for researchers aiming to access large-scale datasets \citep{annas2003hipaa}. Additionally, requirements such as institutional review board (IRB) approvals, legal agreements, and the heterogeneity of data across institutions further complicate data access and collaboration. These barriers often result in datasets that are small, biased, or unrepresentative of the broader population, limiting the development of robust and generalizable models. Techniques such as semi-supervised learning have been proposed to alleviate labeling burdens in medical imaging tasks \citep{zhou2024reducing}.
To address challenges in data availability and interoperability, researchers are increasingly exploring synthetic healthcare data generation as a practical alternative. Synthetic data can replicate the statistical patterns of real-world healthcare datasets, enabling improved data accessibility for model development, benchmarking, and collaborative research \citep{choi2017generating}.

However, ensuring fairness in synthetic data generation is essential, particularly when it comes to accurately representing minority subgroups such as different racial or ethnic populations, age groups, and patients with rare diseases \citep{chen2018my}. Synthetic datasets are increasingly used to train predictive models, validate algorithms, and support clinical decision-making pipelines. Yet, concerns with fairness can arise when generative models replicate or amplify biases in the training data. Biased synthetic data can lead to underdiagnosis, ineffective treatment recommendations, or unequal resource allocation, reinforcing structural inequities in healthcare systems. For instance, \citep{chen2019validity} demonstrated that a GAN trained on the MIMIC-III dataset underrepresented African American patients, resulting in biased outputs and compromising model generalizability.

Recent research has sought to address these concerns by focusing on fairness in synthetic data generation, particularly in healthcare and structured tabular domains \citep{marchesi2023generative, lu2023machine}. \citep{erfanian2024chameleon} showcased this trend with a framework that improves subgroup representativeness. To support standardized assessment, \citep{qian2023synthcity} introduced SYNTHCITY, a benchmarking suite that evaluates synthetic tabular data along the dimensions of representational parity. \citep{giuffre2023harnessing} emphasized the ethical and technical complexities of integrating synthetic data into workflows, while \citep{bhanot2021problem} discussed the role of multi-metric fairness evaluations such as log disparity. 


To support equitable outcomes, high-quality synthetic healthcare data should (1) preserve critical statistical properties and subgroup-level distributions, particularly across key demographic and clinical dimensions; and (2) ensure representation of diverse populations, especially those that are historically underrepresented. Importantly, fairness evaluations should move beyond isolated attributes and consider \textit{intersectional populations}—individuals who belong to multiple demographic subgroups simultaneously (e.g., age, race, and gender). These groups often face compounded disparities and are at a higher risk of being overlooked in data-driven models. Neglecting intersectional subgroups in synthetic data evaluations can result in biased clinical decision-support systems, ultimately compromising the quality of care for vulnerable populations (Examples \ref{intersectional-underrepresentation}).

\begin{minipage}{\columnwidth}
\begin{tcolorbox}[colframe=blue!50!black, colback=purple!10!white, title=Intersectional Bias in Healthcare Data Representation, boxrule=0.95mm, width=0.95\columnwidth, label={intersectional-underrepresentation}]
Breast cancer risk prediction- researchers found that Black women were underrepresented in the training datasets used for machine learning models \citep{park2023evaluation}. 
\vspace{0.5em}
\hrule
\vspace{0.5em}
Older Latino individuals, particularly those with heart failure, were doubly marginalized by both ethnicity and age, leading to suboptimal cardiovascular disease risk predictions using synthetic healthcare data for cardiovascular disease (CVD) risk prediction \citep{rodriguez2014status}.
\vspace{0.5em}
\hrule
\vspace{0.5em}
Young Black women with obesity were underrepresented in synthetic datasets, leading to biased predictions of diabetes risk \citep{cronje2023assessing}.
\end{tcolorbox}
\end{minipage}

Current approaches to measuring fairness and fidelity in synthetic data generation are limited in two key ways. First, most studies often rely on a single generation model, such as HealthGAN \citep{yale2020generation}, which may introduce architectural biases. This narrow focus overlooks the variation in how different models capture complex relationships in real-world clinical data\citep{xu2019modeling}. Second, fairness assessments often rely on only two fairness metrics: the log disparity metric and the time series disparity metric. Although useful, these metrics can produce conflicting results and do not capture the full landscape of biases that may exist on multiple dimensions of the data\citep{bhanot2021problem}. 

\begin{figure}[htbp]
\centering
\includegraphics[width=0.4\textwidth]{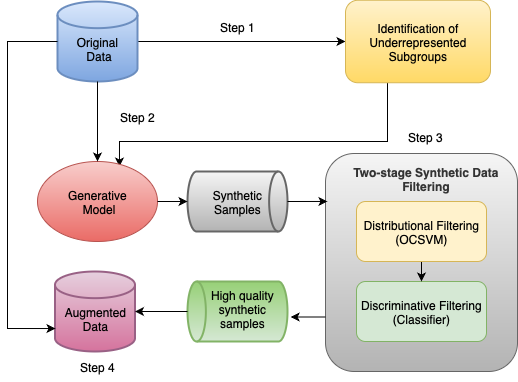}
    \caption{MedEqualizer Workflow}
    \label{fig:workflow}
\end{figure}

Using multiple models and various fairness metrics, our aim is to provide a more comprehensive evaluation of synthetic healthcare data, ultimately promoting its usability, validity, and trustworthiness for research and decision-making purposes. Our contributions are summarized below.
\begin{itemize}
     \item We evaluate multiple generation models for tabular data, MedGAN \citep{xu2018tgan}, HealthGAN \citep{yale2020generation} and Conditional Tabular GAN (CTGAN) \citep{xu2019modeling}, to compare their performance and robustness in producing fair and representative synthetic healthcare data.
 
    \item We develop MedEqualizer, a new framework to mitigate the bias of the synthetic data generation, inspired by the Chameleon framework, as shown in Figure \ref{fig:workflow}.  

    \item Our framework is designed to be model-agnostic and generalizable, enabling its integration with multiple generation models to support diverse healthcare data generation scenarios.

    \item We augment the real dataset with only a small number of additional records, ensuring minimal impact on the underlying data distribution while significantly improving model fairness not just for individual groups but also for intersectional subgroups.
    
\end{itemize}


\section{Preliminaries}

\subsection{MedGAN}
MedGAN (Medical Generative Adversarial Network), introduced by \citep{xu2018tgan}, is one of the earliest generative models developed specifically for synthesizing high-dimensional discrete patient records, such as EHRs. Traditional GANs often struggle with modeling binary or count-based medical codes due to their sparsity, high dimensionality, and lack of continuous structure. MedGAN addresses these challenges by combining an autoencoder with a generative adversarial framework, enabling effective generation of synthetic multi-label healthcare data.
The architecture of MedGAN includes two main components: a pre-trained autoencoder and a GAN. The autoencoder, composed of fully connected layers with non-linear activations, is trained to compress and reconstruct high-dimensional binary vectors representing patient visits. The GAN’s generator then learns to produce latent codes that the decoder can transform into realistic synthetic records, while the discriminator evaluates whether these records resemble real data.

MedGAN introduces three technical innovations to enhance training and data fidelity:
(1) Autoencoder-based decoding: By using a pre-trained decoder to reconstruct discrete outputs, MedGAN bypasses the difficulty of generating binary vectors directly from the generator, thus improving learning efficiency and output quality.
(2) Minibatch averaging: This strategy incorporates average statistics from each batch into the discriminator's input, helping stabilize training and reducing mode collapse by exposing the model to batch-level patterns.
(3) Joint training with reconstruction loss: MedGAN fine-tunes the autoencoder jointly with adversarial training by combining reconstruction loss with GAN loss, promoting output diversity while preserving feature semantics.

\subsection{CTGAN}
CTGAN (Conditional Tabular Generative Adversarial Network), introduced by \citep{xu2019modeling}, represents a significant advancement in synthetic tabular data generation, particularly in healthcare applications. Traditional data generation techniques often struggle with mixed data types, non-Gaussian distributions, and imbalanced categorical variables, common in healthcare datasets. CTGAN addresses these challenges by utilizing a generator and a discriminator network that both employ multi-layer perceptrons with LeakyReLU activations and batch normalization. 
One of CTGAN's core strengths lies in the generator's ability to condition categorical variables, which it encodes through learned embeddings. The discriminator, which evaluates the realism of the generated data, is designed to handle both real and synthetic data, ensuring that the output maintains the statistical properties of the original dataset.

CTGAN's architecture includes three key innovative techniques:  (1) Mode-specific normalization: This method allows the model to transform continuous columns with multi-modal distributions. Using a reversible Gaussian transformation based on identified modes ensures that complex multimodal data distributions are preserved. (2) Conditional generator training: In contrast to traditional GANs, CTGAN incorporates a training-by-sampling method that balances the representation of categorical values across all levels, preventing the model from overfitting certain categories and ensuring fairness. (3)Wasserstein GAN with gradient penalty: This optimization strategy enhances training stability, addressing issues such as mode collapse and the vanishing gradient problem, common in adversarial training. 


\subsection{HealthGAN}
HealthGAN, proposed by \citep{yale2020generation}, represents a specialized approach to generating synthetic health data, with a focus on addressing the complexities of EHRs. EHRs contain diverse data types, such as discrete medical codes, continuous lab values, and temporal information, which must be carefully preserved to maintain clinical validity. HealthGAN builds upon the GAN framework by incorporating domain-specific modifications to handle the heterogeneous nature of health data.

HealthGAN introduces several technical innovations to address healthcare-specific data challenges:
(1) Data normalization for EHRs: The model preprocesses continuous and categorical attributes using tailored scaling and encoding strategies, enabling stable training across heterogeneous data types.
(2) Latent space regularization: By integrating reconstruction and adversarial losses during training, HealthGAN ensures that the latent space encodes meaningful representations that capture real-world data variability without memorizing sensitive details.
(3) Encoder-assisted generation: The model uses a trained encoder to improve sampling diversity and realism, supporting conditional sampling and facilitating evaluation of generated samples.

\subsection{Chameleon}
Chameleon \citep{erfanian2024chameleon} is a system for fairness-aware multi-modal data augmentation that leverages foundation models to enhance the representation of minority groups. Below, we discuss three (Combination Selection, Foundation Model Generation, and Rejection Sampling) of the four major components from Chameleon that we adapted in this paper. While Chameleon also employs a Guide Tuple Selection mechanism, which uses a selective masking of features at a time, our approach differs by considering combinations of demographic features (gender, age, and race) in the datasets jointly. This allows for a more comprehensive analysis of subgroup-level imbalances, especially in intersectional populations that are often underrepresented in clinical datasets. 

\subsubsection{Combination Selection}
The process begins by identifying uncovered patterns in the dataset using the concept of data coverage. Given a dataset $D$ and a coverage threshold $\tau$, a subgroup $g$ is considered uncovered if $|g \cap D| < \tau$. These uncovered subgroups are often underrepresented in the original data, which may lead to biases in the final model. The system identifies Maximal Uncovered Patterns (MUPs) representing these underrepresented demographic or clinical groups, ensuring that synthetic data generation addresses any imbalances.

To address these coverage issues, the Combination-Selection algorithm identifies the minimal set of synthetic tuples required. This algorithm employs a greedy approximation approach, iteratively selecting combinations of data points that cover the maximum number of remaining uncovered patterns. This approach minimizes the total number of synthetic samples required, ensuring that the resulting synthetic data maintains a balanced representation of all subgroups without overproducing synthetic samples for already well-represented patterns.

\subsubsection{Foundation Model Generation}
Chameleon selects a selective featured-based guide tuple and corresponding mask to pass to a foundation model (e.g., DALL·E 2 \citep{dalle2_openai}). The mask delineates the foreground subject, indicating which parts of the tuple are to be regenerated while preserving the background context. The combination of the guide tuple and the generated content results in a synthetic sample that is realistic and conforms to the original data distribution. 

\subsubsection{Rejection Sampling}
After the foundation model generates synthetic data, the generated tuples undergo two quality assurance tests to ensure their validity: (1) Data Distribution Test: This test employs a one-class support vector machine (SVM) \citep{scholkopf1999novelty} in the embedding space to verify whether the generated samples adhere to the underlying data distribution. 
(2) Quality Evaluation Test: This test employs a hypothesis testing to determine if the generated tuples appear realistic to human evaluators. 

The above quality assurance steps are important to ensure the integrity of the synthetic data for high-stakes applications such as healthcare, where the data should be valid and representative to support fair and accurate decision-making. In this paper, we employ a two-stage synthetic data filtering approach inspired by Chameleon \citep{erfanian2024chameleon} to select synthetic data samples for augmenting underrepresented subgroups. We describe the process further in detail in Section \ref{sec:fairness_aware_aug} and Figure \ref{fig:workflow}.

Our work builds upon the Chameleon framework \citep{erfanian2024chameleon}, extending it for tabular data in multiple aspects to address fairness in synthetic data generation more effectively. While Chameleon focuses on improving subgroup coverage through a two-stage filtering and augmentation process, our approach adapts this approach explicitly for \textit{demographically stratified health datasets and intersectional population}. Additionally, we apply the framework across three generative models (MedGAN, CTGAN and HealthGAN), offering model-agnostic insights into how fairness-aware augmentation influences representational equity in healthcare.

\section{Dataset}
\subsection{The MIMIC-III Database}
This study utilizes the Medical Information Mart for Intensive Care III (MIMIC-III) database, a large, publicly available database comprising de-identified electronic health records of 38,623 distinct patients admitted to the Beth Israel Deaconess Medical Center in Boston, Massachusetts between 2001 and 2012. MIMIC-III has been widely used in clinical research due to its breadth and depth, offering detailed information across clinical variables such as diagnoses, medications, procedures, lab measurements, and outcomes. 

For our fairness-aware data augmentation and debiasing analysis, we focus specifically on the demographic attributes and the mortality outcomes recorded in this database. These variables are important for evaluating the presence of disparities and biases in machine learning models trained for clinical decision making and predictive tasks, especially those relevant to the context of ICU mortality. 
A notable characteristic of MIMIC-III for fairness research is its demographic composition, which presents opportunities as well as challenges for fairness evaluation. The database comprises data from adult patients with a median age of 66 years, of whom 56.6\% are male and an overall in-hospital mortality rate of 40.57\%. The racial distribution includes approximately 71.32\% White, 7.64\% Black, 3.26\% Hispanic, 2.37\% Asian, and 15.4\% patients categorized as other or unknown race/ethnicity. While this demographic composition allows for meaningful analysis of potential disparities across demographic groups, the substantial imbalance, especially among racial and ethnic subgroups, poses challenges for bias detection and correction, emphasizing the need for targeted augmentation techniques.

Access to MIMIC-III requires the completion of a data use agreement and training in human subjects research protection. We obtained access to the database through the PhysioNet repository \cite{mimic3} after satisfying the prerequisites.

\subsection{Cohort Selection}
To select a cohort appropriate for fairness analysis, we applied a set of inclusion and exclusion criteria to the entire MIMIC-III dataset. We included adult patients ($\geq$18 years) admitted to the ICU with documented demographic information (including age, gender, and race/ethnicity). 
To minimize the issue of missing data in fairness-critical features, we excluded patients with missing demographic information or mortality outcomes. Additionally, we excluded the small number of patients whose race/ethnicity was recorded as ``Unable to Obtain'' or left blank, as these categories provided insufficient information for our fairness analysis framework. 
After applying these filters, our final cohort consisted of 
gender, race, age, mortality rate, admission location,
admission type, insurance, disease, first care unit,
and last care unit, maintaining sufficient representation across key demographic groups to enable meaningful bias detection and mitigation. Table~\ref{tab_data} presents the detailed distribution of demographic characteristics in our analytical cohort, highlighting the composition of subgroups relevant to fairness evaluation.

\begin{table}[h]
\centering
\caption{Distribution of Age, Race, and Gender}
\begin{tabular}{lll}
\hline
Feature & Category & Percentage (\%) \\
\hline
\textbf{Age} & $\leq$ 45 & 15.45 \\
             & 45-65 & 33.61 \\
             & 66-80 & 30.85 \\
             & 81+   & 20.09 \\
\hline
\textbf{Race} & Asian   & 2.37 \\
              & Black   & 7.64 \\
              & White   & 71.23 \\
              & Other   & 3.53 \\
              & Unknown & 15.22 \\
\hline
\textbf{Gender} & Male & 56.6 \\
                & Female & 43.4 \\
\hline
\label{tab_data}
\end{tabular}
\end{table}

\subsection{Fairness-Critical Features}
Similar to prior works, such as \citep{huang2022evaluation}, we extracted a focused set of features from the MIMIC-III dataset that are specifically related to fairness concerns in clinical prediction. These include protected attributes as well as context variables that influence clinical treatment/care decisions and outcomes:
\begin{enumerate}
\item \textbf{Protected Attributes}: We identified \textit{age}, \textit{gender}, and  \textit{race} as protected attributes with potential for bias, consistent with frameworks proposed by \citep{pfohl2019creating}. Age was treated as both a continuous variable and a discretized categorical variable. For categorical analysis, we grouped patients into clinically relevant categories ($\leq$45, 46–65, 66–80, and +81 years). Gender was recorded as a binary variable (male/female) in the original dataset. We acknowledge this as a limitation, as the dataset does not represent non-binary identities. Race/ethnicity was categorized as documented in MIMIC-III (White, Black, Asian, Other, and Unknown). 

\item \textbf{Mortality Outcomes}: We used 30-day mortality as a binary variable categorized as died and alive in our analysis.

\item \textbf{Insurance Status}: We included insurance type (Medicare, Medicaid, Private, Self Pay, and Government) as a proxy for socioeconomic status, which may influence access to care and provider decisions. This variable can potentially be an important contextual factor in fairness evaluations.

\item \textbf{Admission Information}: We incorporated admission type (elective, urgent, emergency, and newborn) and admission location (emergency room admit, hospital or external transfers, physician or clinic referrals, skilled nursing facilities, other healthcare facilities, HMO referrals, and intra-facility transfers) to account for potential selection biases in care pathways.

\item \textbf{Disease}:
Disease type was categorized using ICD-9 diagnosis codes by grouping patients into one of four categories: any malignancy, congestive heart failure, both (if both conditions were present), or other (if neither condition was detected). This categorical variable could help reveal potential disparities across disease groups.

\item \textbf{First care unit and last care unit}: We included first care unit and last care unit (TSICU, MICU, CCU, SICU, CSRU, and NICU) as this variable may provide context for fairness evaluations.

\end{enumerate}

\section{MedEqualizer Workflow}

In this section, we outline the MedEqualizer workflow. First, we examine subgroup representation disparities in synthetic data generated by different models. Then, we present a debiasing approach that uses targeted synthetic data augmentation to reduce disparities, promote subgroup coverage and improve fair representation across protected subgroups.

\subsection{Logarithmic Disparity Metric}
To quantify representational bias in synthetic data, we adopt the \textit{Logarithmic Disparity (Log Disparity)} metric \citep{qi2021quantifying}, a fairness metric used to evaluate the resemblance between synthetic and real healthcare data, specifically focusing on the representation of minority subgroups. It measures the logarithmic difference between the proportions of a specific attribute value (e.g., a particular race or gender or both) in the synthetic and real datasets. 
Formally, for a given attribute value or subgroup, the logarithmic disparity metric can be defined as:
\[ \text{Log Disparity} = \log\left(\frac{p_s}{p_r}\right) \]
where $p_s$ and $p_r$ are the proportions of the specific attribute value in the synthetic and real datasets, respectively.
We apply this metric across multiple protected attributes and their intersections to systematically analyze disparities introduced by different synthetic data generation models.

\subsection{Subgroup Representativeness Analysis}
To evaluate fairness in the generated synthetic data, we evaluate the representativeness of different demographic subgroups using the logarithmic disparity metric defined earlier. This analysis quantifies how well each demographic subgroup is preserved with the synthetic data generation compared to the real data, allowing us to identify both the underrepresented and the overrepresented groups/subgroups. 

We visualize the representativeness of each subgroup using hierarchical sunburst diagrams, which effectively conveys multivariate disparities across combinations of protected attributes. In a sunburst diagram, each concentric ring corresponds to a different demographic attribute (e.g., mortality status, race/ethnicity, age, and gender), with inner rings representing broader categories and outer rings capturing finer subgroup divisions. The color coding in the sunburst diagrams reflects the degree of subgroup representativeness, based on the logarithmic disparity metric computed for that subgroup. Specifically, 
\begin{itemize}
    \item \textcolor{darkblue}{\rule{1ex}{1ex}} \textbf {Dark blue}
    indicates highly overrepresented subgroups
    
    (log disparity value > –log(0.8))
    \item \textcolor{blue}{\rule{1ex}{1ex}} \textbf{Blue} indicates overrepresented subgroups 
    
    (between –log(0.9) and –log(0.8))
    \item \textcolor{teal}{\rule{1ex}{1ex}} \textbf{Teal} represents adequately represented subgroups 
    
    (between log(0.9) and –log(0.9)
    \item \textcolor{lightorange}{\rule{1ex}{1ex}} \textbf{Light orange} indicates underrepresented subgroups 
    
    (between log(0.8) and log(0.9))
    \item \textcolor{orange}{\rule{1ex}{1ex}}  \textbf{Orange} indicates highly underrepresented subgroups 
    
    (less than log(0.8))
\end{itemize}
This hierarchical and color-coded structure allows intuitive identification of biases across multivariate combinations of protected attributes \citep{bhanot2021problem}. This visual and hierarchical format offers an intuitive yet rigorous platform to visualize disparities across intersectional demographic subgroups. 

\subsection{Fairness-Aware Data Augmentation}
\label{sec:fairness_aware_aug}
To mitigate subgroup imbalanced representation and promote fairness in downstream tasks, we adapted key ideas from the Chameleon framework \citep{erfanian2024chameleon} to develop a targeted augmentation strategy for underrepresented demographic groups in our dataset. Figure \ref{fig:workflow} illustrates the overall workflow of our approach. Next, we describe each step of the workflow.

\subsubsection{Identification of Underrepresented Subgroups} (Figure \ref{fig:workflow} Step 1)
We identify minority combinations of gender, race, and age based on a frequency threshold $\tau$ in real data. Any demographic combination with a count below $\tau$ was flagged as a minority subgroup requiring augmentation. We used a threshold of $\tau = 150$ to identify underrepresented subgroups in the real dataset, meaning any subgroup with fewer than 150 samples was selected for augmentation. We empirically tested other thresholds, including \( \tau = 100 \) and \( \tau = 200 \), but found that \( \tau = 150 \) provided the best balance between subgroup targeting and overall data representation.


\subsubsection{Model-Specific Generation Strategies} (Figure \ref{fig:workflow} Step 2)
For synthetic data generation, we employed three generative models: CTGAN, HealthGAN and MedGAN, customizing the augmentation pipeline for each:

\begin{itemize}
    \item CTGAN: We trained a conditional generative model on the entire dataset and generated synthetic samples, conditioning the model on each underrepresented subgroup to ensure targeted generation. This model learns the joint distribution across all demographic groups. This approach helps retain the overall data structure while increasing the minority group representation. 
    \item HealthGAN and MedGAN: Unlike CTGAN, these models do not support conditional generation. Hence, we trained a separate model for each underrepresented subgroup using only real samples corresponding to that subgroup. This subgroup-specific training allowed the model to learn and capture the unique characteristics of each demographic combination. 
\end{itemize}  

\subsubsection{Two-Stage Synthetic Data Filtering} (Figure \ref{fig:workflow} Step 3)
For all models, we generate synthetic samples in batches of size $N$ for each target demographic combination. This batch-based approach allows us to implement quality control through a two-stage filtering process:

\begin{itemize}
    \item Distributional Filtering: We train a One-Class Support Vector Machine (OCSVM) on the real data distribution.  The model learns the boundary that encompasses the majority of the real data points in the feature space. For each generated batch, we apply this OCSVM to filter out samples that fall outside the learned boundary, retaining only those match real data.
    \item Discriminative Filtering: We trained a logistic regression classifier to distinguish between real samples and the generated samples that passed the first filter. We compute the Area Under the ROC Curve (AUC) for this classifier and only accept a batch if the AUC score falls below a predefined threshold $\alpha$. A low AUC indicates that the classifier struggles to distinguish between real and synthetic samples, suggesting high similarity between the two.
\end{itemize} 
This combined two-stage filtering strategy, distributional and discriminative, helped ensure that only high-fidelity synthetic generated data were included. 

\subsubsection{Data Augmentation} (Figure \ref{fig:workflow} Step 4)
After generating synthetic records that pass both tests for all underrepresented subgroups, we integrate them into the real dataset. As a result, the augmented data improves demographic balance while preserving the essential characteristics of the original dataset. This threshold-based approach ensures that augmentation efforts are both focused and minimal, intervening only where subgroup disparities are more pronounced. 

\subsection{MedEqualizer Algorithm}
Algorithm 1 shows the step-by-step procedure of the MedEqualizer, which identifies underrepresented subgroups and generates synthetic samples to enhance fairness in the dataset. Table \ref{tab:algorithm_vars} provides a detailed description of all variables used in the algorithm. The algorithm begins by identifying underrepresented subgroups $\mathcal{C}$ in the real data $\mathcal{D}$ based on a predefined coverage threshold $\tau$ (line 1). A subgroup $(g, r, a)$ defined by gender, race, and age is considered underrepresented if its size is less than $\tau$. For each underrepresented subgroup $\mathcal{D}_{g,r,a}$, the algorithm calculates the required number of synthetic samples, called the gap, computed as $\tau - |\mathcal{D}_{g,r,a}|$. Synthetic sample batches $S$ of size $N$ ($N < $ gap) are then generated iteratively (line 6). If the generator $G$ is CTGAN, samples are generated conditionally; if $G$ is MedGAN or HealthGAN, a separate model is first trained on $\mathcal{D}_{g,r,a}$ and then used to generate samples. To ensure the quality of the generated data, a two-stage filtering process is applied for each batch $S$:
\begin{itemize} 
\item A One-Class SVM is trained on $\mathcal{D}$ and applied to $S$ to retain only distributionally similar samples, resulting in $S_{\text{valid}}$ (line 7)
\item A classifier is then trained to distinguish between real samples $\mathcal{D}_{g,r,a}$ and synthetic samples $S_{\text{valid}}$. The batch is accepted into $A_{g,r,a}$ only if the classifier’s AUC is less than or equal to $\alpha$ (lines 8--11).
\end{itemize}
This process repeats until the size of $A_{g,r,a}$ meets the gap, ensuring that the subgroup reaches the target threshold $\tau$. After each subgroup is processed, the valid samples $A_{g,r,a}$ are added to the overall collection $A$ (line 12), which accumulates validated synthetic samples across all subgroups. Finally, the algorithm returns the augmented dataset $\mathcal{D}_{\text{aug}} = \mathcal{D} \cup A$ (line 15), combining the real data with all accepted synthetic samples to achieve better demographic balance.

\begin{algorithm}[t]
\caption{MedEqualizer Algorithm}
\begin{algorithmic}[1]
\label{alg:fairness}
\REQUIRE $\mathcal{D}$, $\tau$, $G$, $N$, $\alpha$
\STATE Identify underrepresented combinations $\mathcal{C}$ where frequency $< \tau$
\FOR{each subgroup $(g, r, a) \in \mathcal{C}$}
    \STATE Filter $\mathcal{D}_{g,r,a} \subset \mathcal{D}$
    \STATE $A \leftarrow \emptyset$
    \WHILE{|A| $<$ gap}
        \STATE $S \leftarrow \texttt{sample}(G, (g, r, a), N)$
        \STATE Apply OCSVM to $S$ to get $S_{\text{valid}}$
        \STATE Train classifier on $\mathcal{D}_{g,r,a}$ and $S_{\text{valid}}$
        \IF{Classifier AUC $\leq \alpha$}
            \STATE Add $S_{\text{valid}}$ to ${A}_{g,r,a}$
        \ENDIF
        \STATE Add ${A}_{g,r,a}$ to $A$
    \ENDWHILE
\ENDFOR
\RETURN $\mathcal{D}_{\text{aug}} = \mathcal{D} \cup$ $A$
\end{algorithmic}
\end{algorithm}

\begin{table}[h]
\centering
\caption{Notations of MedEqualizer Algorithm ~\ref{alg:fairness}}
\label{tab:algorithm_vars}
\small
\begin{tabular}{lp{7cm}}
\hline
\textbf{Symbol} & \textbf{Description} \\
\hline
$\mathcal{D}$ & Real dataset \\
$\mathcal{C}$ & Set of underrepresented subgroup combinations \\
$\tau$ & Coverage threshold for identifying underrepresented subgroups \\
$G$ & Data generator (MedGAN, HealthGAN, or MedGAN) \\
$N$ & Batch size for synthetic sample generation \\
$\alpha$ & Classifier acceptance threshold (AUC) \\
$(g, r, a)$ & Gender, race, and age combination defining a subgroup \\
$\mathcal{D}_{g,r,a}$ & Subset of $\mathcal{D}$ matching a given subgroup $(g, r, a)$\\
$S$ & Generated synthetic sample batch \\
$S_{\text{valid}}$ & Subset of $S$ passing the OCSVM distributional filter \\
$A_{g,r,a}$ & Set of valid synthetic samples accepted for subgroup $(g,r,a)$ \\
$A$ & Aggregated set of all accepted synthetic samples across subgroups \\
$gap$ & Number of synthetic samples needed; computed as $\tau - |\mathcal{D}_{g,r,a}|$ \\
$\mathcal{D}_{\text{aug}}$ & Final augmented dataset \\
\hline
\end{tabular}
\end{table}

\section{Results and Discussion}
\begin{figure}[htbp]
    \centering
    \begin{subfigure}[b]{0.35\textwidth}
        \includegraphics[width=\textwidth]{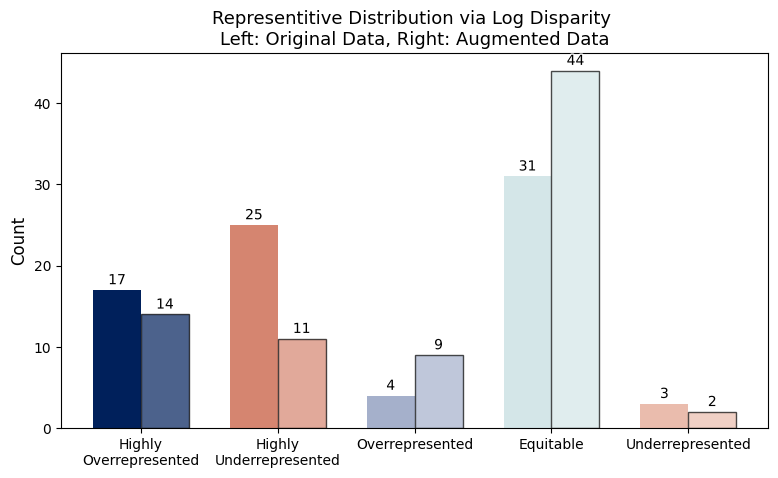}
        \caption{MedGAN}
        \label{fig:sub-medgan}
    \end{subfigure}
    \hfill
    \begin{subfigure}[b]{0.35\textwidth}
        \includegraphics[width=\textwidth]{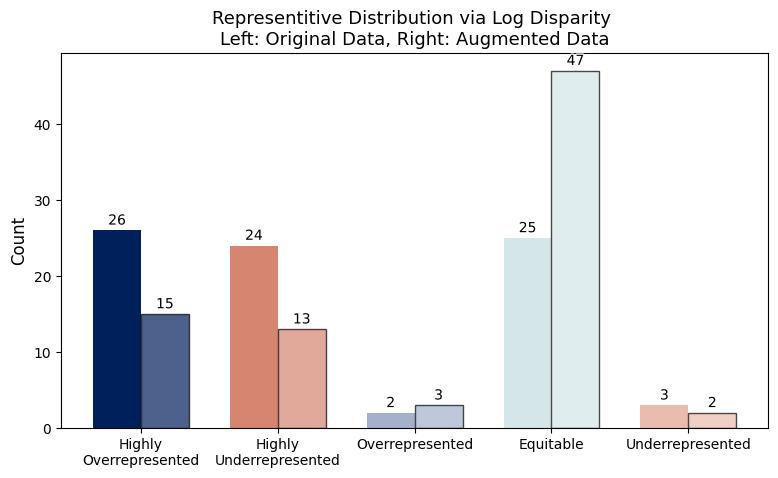}
        \caption{HealthGAN}
        \label{fig:sub-healthgan}
    \end{subfigure}
    \begin{subfigure}[b]{0.35\textwidth}
        \includegraphics[width=\textwidth]{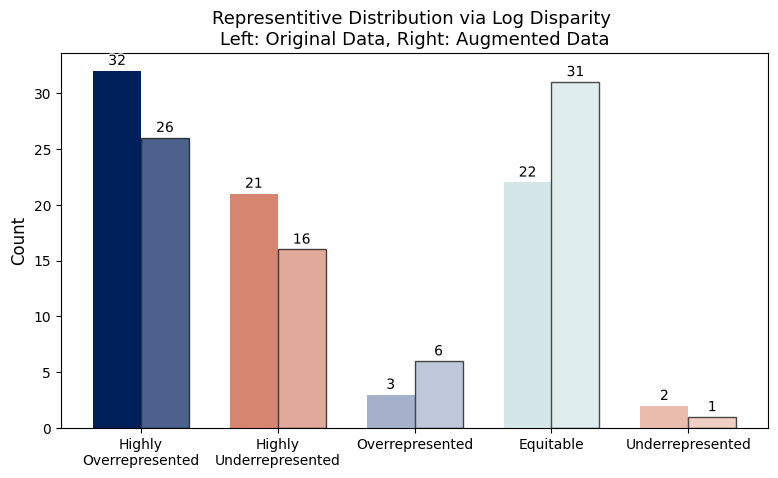}
        \caption{CTGAN}
        \label{fig:sub-ctgan}
    \end{subfigure}
    \caption{Histograms showing the representation of all demographic subgroup combinations (age, race, and gender) in synthetic data compared to real data. For each model: (a) MedGAN, (b) HealthGAN, and (c) CTGAN. The left bars correspond to data generated without augmentation, and the right bars correspond to data augmented using MedEqualizer.}
    \label{fig:combined_hist}
\end{figure}


\begin{figure*}[htbp]
    \centering
    \begin{subfigure}[b]{0.45\textwidth}
        \includegraphics[width=\textwidth]{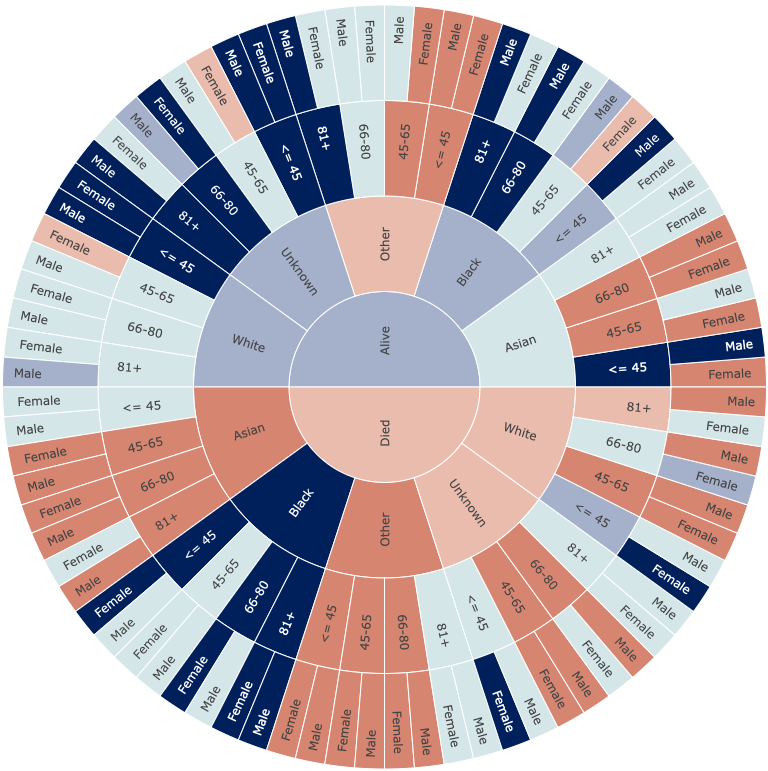}
        \caption{before data augmentation}
        \label{fig:medgan-before}
    \end{subfigure}
    \hfill
    \begin{subfigure}[b]{0.45\textwidth}
        \includegraphics[width=\textwidth]{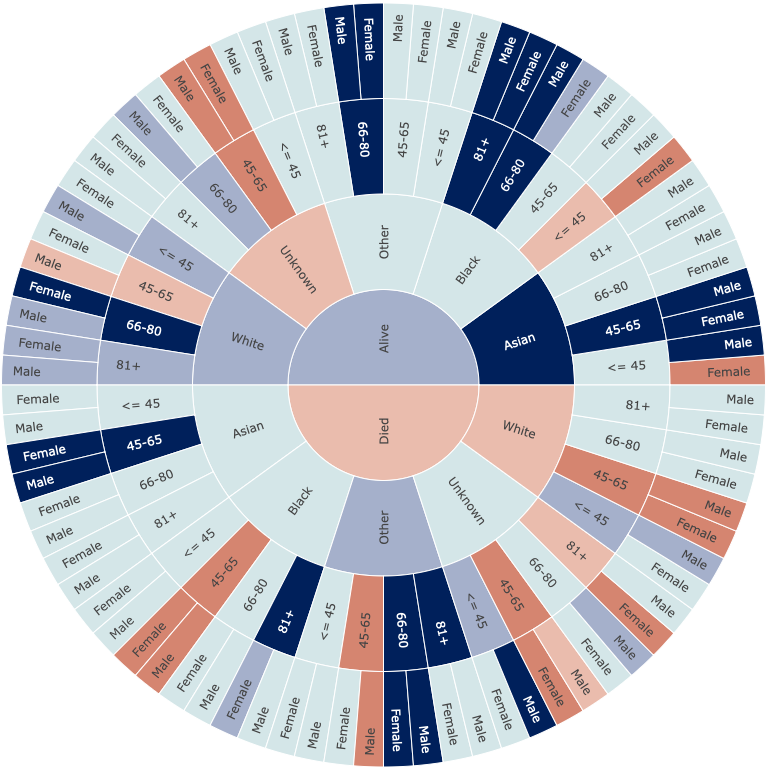}
        \caption{after data augmentation}
        \label{fig:medgan-after}
    \end{subfigure}
    \caption{Subgroup representativeness in MIMIC-III synthetic data generated by MedGAN (a) before and (b) after data augmentation.}
    \label{fig:medgan}
\end{figure*}

\begin{figure*}[htbp]
    \centering
    \begin{subfigure}[b]{0.45\textwidth}
        \includegraphics[width=\textwidth]{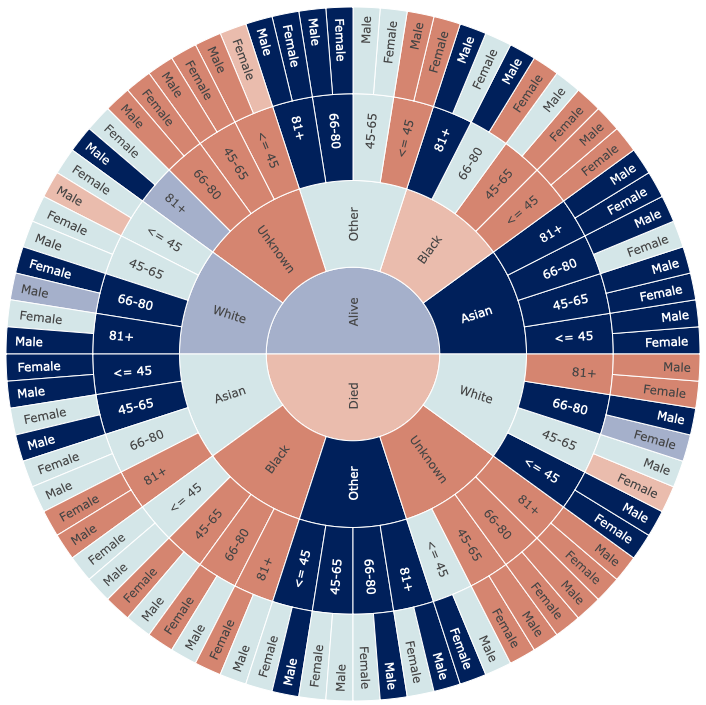}
        \caption{before data augmentation}
        \label{fig:healthgan-before}
    \end{subfigure}
    \hfill
    \begin{subfigure}[b]{0.45\textwidth}
        \includegraphics[width=\textwidth]{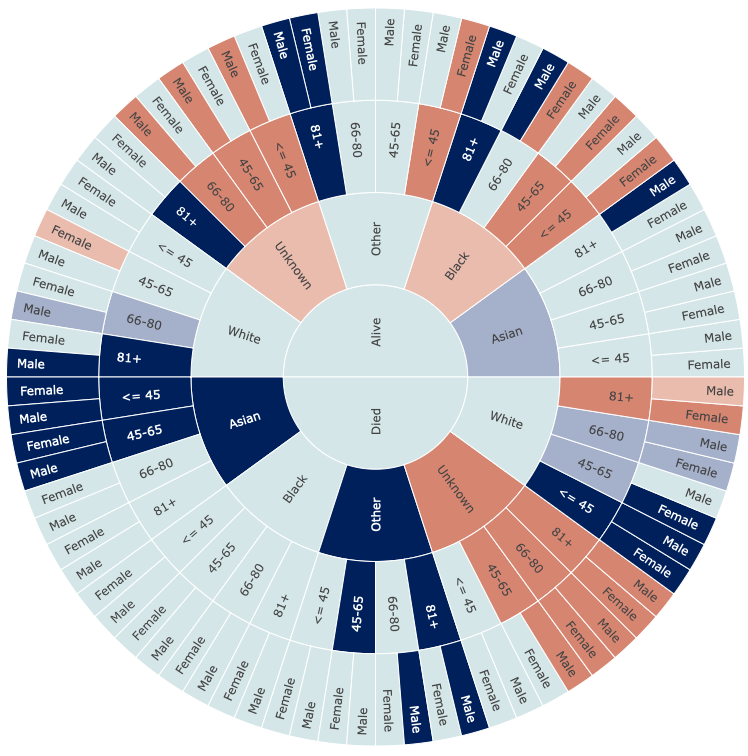}
        \caption{after data augmentation}
        \label{fig:healthgan-after}
    \end{subfigure}
    \caption{Subgroup representativeness in MIMIC-III synthetic data generated by HealthGAN (a) before and (b) after data augmentation.}
    \label{fig:healthgan}
\end{figure*}

\begin{figure*}[htbp]
    \centering
    \begin{subfigure}[b]{0.45\textwidth}
        \includegraphics[width=\textwidth]{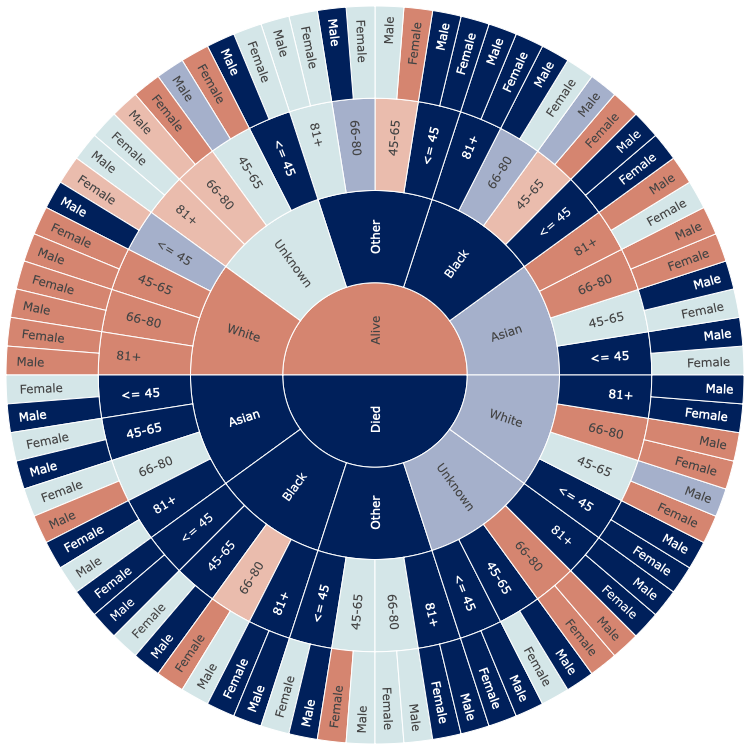}
        \caption{before data augmentation}
        \label{fig:ctgan-before}
    \end{subfigure}
    \hfill
    \begin{subfigure}[b]{0.45\textwidth}
        \includegraphics[width=\textwidth]{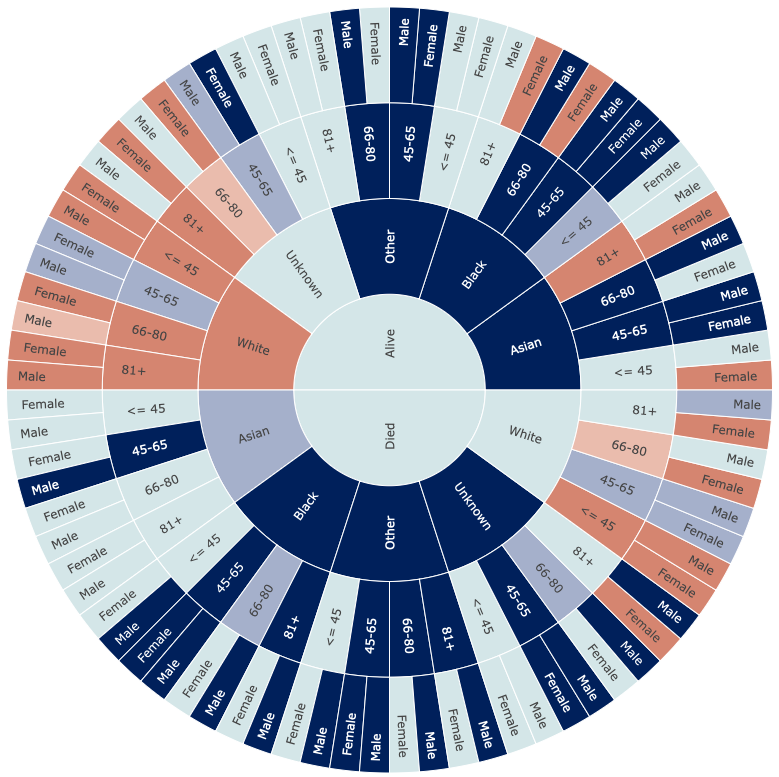}
        \caption{after data augmentation}
        \label{fig:ctgan-after}
    \end{subfigure}
    \caption{Subgroup representativeness in MIMIC-III synthetic data generated by CTGAN (a) before and (b) after data augmentation.}
    \label{fig:ctgan}
\end{figure*}

We generated synthetic data from the MIMIC-III dataset using three different GAN-based models for tabular data: MedGAN, HealthGAN, and CTGAN. In this section, we present and analyze the subgroup representativeness of synthetic data generated by these models compared to the real MIMIC-III data, as visualized in the sunburst diagrams (Figures (a) in \ref{fig:medgan}, \ref{fig:healthgan} and \ref{fig:ctgan}). These diagrams summarize log disparity values across multivariate demographic subgroups defined by mortality, race, age, and gender. 
Each model generates different proportions of demographic subgroups, indicating notable differences in representational fairness. 
For example, CTGAN (Figure ~\ref{fig:ctgan}a) exhibits a substantial underrepresentation of the \textit{Alive–White} subgroups, especially among older female patients. In contrast, the same subgroups in HealthGAN (Figure ~\ref{fig:healthgan}a) are almost adequately represented and in MedGAN (Figure ~\ref{fig:medgan}a  are either adequately represented or overrepresented. Conversely, the \textit{Died-Black male} subgroup is highly overrepresented in CTGAN-generated synthetic data, while HealthGAN and MedGAN maintain a more proportionate representation of this subgroups. Overall, MedGAN and HealthGAN demonstrate a more balanced demographic distribution between subgroup combinations, suggesting superior capability in mitigating representation bias across intersectional attributes compared to CTGAN, which shows more pronounced disparities. 

We also applied the MedEqualizer framework to augment the MIMIC-III dataset and generated synthetic data from the augmented version using all three models. 
These augmented datasets were then fed into the same GAN model used during augmentation to generate the post-augmentation synthetic data. We analyze the resulting sunburst diagrams to evaluate the impact of MedEqualizer on subgroup representativeness comared to the real MIMIC-III data, as shown in Figures~(b) in \ref{fig:medgan}, \ref{fig:healthgan}, and \ref{fig:ctgan}.


Comparing panels (b) with (a) in Figures~\ref{fig:medgan}, \ref{fig:healthgan}, and \ref{fig:ctgan}, we observe substantial improvements in subgroup balance after data augmentation, particularly in synthetic data generated by MedGAN and HealthGAN.
In the MedGAN-generated data, many subgroups show improved representation. For example, the \textit{Died–Asian} subgroups aged \textit{66–80} and \textit{81+} transition from being highly underrepresented (Figure~\ref{fig:medgan}a) to adequately represented (Figure~\ref{fig:medgan}b). Similarly, the highly overrepresented \textit{Died–Black–Female} subgroups aged \textit{$\leq$45} and \textit{66–80} become adequately represented.

In the HealthGAN-generated data, the \textit{Died–Black} subgroups across all age and gender categories shift from being highly underrepresented (Figure~\ref{fig:healthgan}a) to adequately represented (Figure~\ref{fig:healthgan}b). Notably, the \textit{Alive–Asian} subgroup—patients aged \textit{$\leq$80}, both male and female—also become adequately represented. Additionally, the \textit{Alive–White–Female} subgroup aged \textit{66–80} transitions from highly overrepresented to moderately overrepresented.

In the CTGAN-generated data, the \textit{Died–Asian} subgroups across nearly all age groups for both males and females shift from being overrepresented (Figure~\ref{fig:ctgan}a) to adequately represented (Figure~\ref{fig:ctgan}b). The \textit{Alive–Black–Male} subgroup aged \textit{81+} improves similarly. Furthermore, the \textit{Alive–Other} subgroups aged \textit{$\leq$45} and \textit{81+}, across both genders, also move toward adequate representation.

To further evaluate the effect of MedEqualizer, we present histograms summarizing the distribution of all combinations of age, race, and gender in the original (real) and augmented synthetic datasets (Figure~\ref{fig:combined_hist}). These histograms classify each subgroup into one of five representational tiers—\textit{highly overrepresented, overrepresented, equitably represented, underrepresented,} and \textit{highly underrepresented}—based on the logarithmic disparity metric. By comparing the distribution of these tiers before and after augmentation for MedGAN, HealthGAN, and CTGAN, we provide a quantitative view of each model’s ability to capture demographic diversity and balance, as discussed below.

MedGAN showed improved subgroup balance following MedEqualizer augmentation. The number of highly overrepresented combinations decreased from 17 to 14, while highly underrepresented subgroups dropped by 13, from 25 to 12. Additionally, the number of equitably represented combinations increased from 31 to 44 after augmentation.

HealthGAN also showed balance using MedEqualizer. The number of highly overrepresented subgroups declined from 26 to 15, and highly underrepresented subgroups decreased from 24 to 13. Meanwhile, the number of equitably represented groups increased substantially, from 25 to 47. This improvement demonstrates that subgroup amplification and omission were mitigated, and synthetic data quality improved with respect to representational fairness.

For CTGAN, MedEqualizer led to a meaningful reduction in subgroup-level imbalance. Specifically, the number of highly overrepresented subgroups decreased from 32 to 26, suggesting that synthetic data generation became less skewed towards certain demographic clusters. Likewise, the number of highly underrepresented subgroups declined from 21 to 16, indicating increased coverage of marginalized or rare subgroup combinations. This indicates a transition towards evenly distributed subgroup representation in the generated synthetic data.

Notably, both models showed a substantial increase in the number of equitably represented subgroups. CTGAN saw an increase from 22 to 31 equitable subgroups, while HealthGAN improved from 25 to 38. This increase reflects a strong improvement in fairness and fidelity of demographic representation. In addition, both models showed a modest increase in moderately overrepresented subgroups and a decrease in moderately underrepresented ones, suggesting a rebalancing effect where subgroup frequencies can adjust towards a central, equitable region. 

Collectively, these results show strong empirical evidence that our fairness-aware data augmentation strategy not only mitigates extreme disparities in subgroup representation but also makes the overall distribution towards a more balanced and representative state. These results have the potential to benefit downstream clinical modeling applications that depend on unbiased/generalized training datasets.


\section{Ethical Considerations}
The proposed MedEqualizer addresses fairness in synthetic healthcare data generation, especially the underrepresented intersectional populations in generative models such as MedGAN, CTGAN and HealthGAN. We recognize that imbalanced synthetic datasets can perpetuate existing biases in real-world healthcare applications. The goal of our proposed fairness aware synthetic data augmentation framework, \textit{MedEqualizer}, is to mitigate these issues by ensuring that the generated synthetic data better mimics the diverse and marginalized subgroups.


However, augmenting real-world healthcare datasets such as MIMIC-III introduces important considerations around data authenticity and the potential overrepresentation of rare patient profiles, particularly those with uncommon conditions. An additional challenge is the risk of relying on synthetic data for clinical decision-making without adequate clinical validation. We explicitly position MedEqualizer as a tool for fairness evaluation and modeling purposes, not for direct clinical inference. MedEqualizer is intended to promote inclusivity and support equitable algorithmic design. We advocate that any future applications involving MedEqualizer include collaboration with community stakeholders and clinical experts to ensure context-appropriate and responsible use.

\section{Conclusion}
In this study, we introduce MedEqualizer, a fairness-aware data augmentation strategy to improve the demographic representativeness of synthetic healthcare data generated by generative models, especially CTGAN and HealthGAN. Using the logarithmic disparity metric, we systematically evaluate the subgroup-level representational biases across combinations of race, gender, and age in populations. Our analysis shows that both models exhibit significant disparities in subgroup representation, often underrepresenting minority populations important for equitable downstream clinical tasks.  

To address these biases, MedEqualizer improves underrepresented subgroups using conditional generation and discriminative filtering. Sunburst visualizations and histogram analyses showed a reduction in both highly over- and underrepresented groups, with HealthGAN demonstrating more balanced generation post-augmentation. These findings highlight the importance of fairness-aware strategies in synthetic data generation. By improving subgroup and intersectional representation, such methods can support the development of more equitable and reliable healthcare models. 

Our efforts to minimize data augmentation while enhancing model fairness ensure that these datasets can better serve diverse populations, foster fairer outcomes in healthcare applications, and ultimately support the development of more inclusive and trustworthy healthcare models. In the future, we will explore additional attributes, datasets, generative architectures and assess downstream impacts on model performance.

 \section{Acknowledgments}

We thank the PhysioNet team and the MIT Laboratory for Computational Physiology for providing access to the MIMIC-III clinical database used in this study. This work was supported by National Institutes of Health grants 5P30EY032857 and utilized computing resources from the Ohio Supercomputer Center.



\bibliographystyle{ACM-Reference-Format}
\bibliography{sample-base}

\appendix

\end{document}